\documentclass[conference]{IEEEtran}
\IEEEoverridecommandlockouts
\usepackage{cite}
\usepackage{amsmath,amssymb,amsfonts}
\usepackage{algorithmic}
\usepackage{graphicx}
\usepackage{textcomp}
\usepackage[hidelinks]{hyperref}
\usepackage{xcolor}
\usepackage{lipsum}
\usepackage{booktabs}
\usepackage{url}
\def\BibTeX{{\rm B\kern-.05em{\sc i\kern-.025em b}\kern-.08em
    T\kern-.1667em\lower.7ex\hbox{E}\kern-.125emX}}

\newcommand{ \bm }[1]{ \mbox{\bf {#1}} }

\newcommand{\bz}{\bm{z}}

\newcommand{\bx}{\bm{x}}

\newcommand{\bX}{\bm{X}}

\newcommand{\bZ}{\bm{Z}}

\begin{document}

\title{Concept Saliency Maps to Visualize Relevant Features in Deep Generative Models\\
{}	
\thanks{This study was partially supported by Narodowe Centrum Nauki grant no. 2016/23/D/ST6/03613. The GPU used for this research was donated by the NVIDIA Corporation. }
}

\author{\IEEEauthorblockN{Lennart Brocki}
	\IEEEauthorblockA{\textit{Institute of Theoretical Physics,} University of Wroclaw \\
		\textit{Institute of Informatics,} University of Warsaw\\
		lennart.brocki@uwr.edu.pl} 
	\and
	\IEEEauthorblockN{Neo Christopher Chung}
	\IEEEauthorblockA{\textit{Institute of Informatics,} University of Warsaw \\ 
	\textit{NIH BD2K Center of Excellence for Biomedical Computing,} \\
	 University of California, Los Angeles \\
		nchchung@gmail.com}
}

\maketitle

\begin{abstract}
Evaluating, explaining, and visualizing high-level concepts in generative models, such as variational autoencoders (VAEs), is challenging in part due to a lack of known prediction classes that are required to generate saliency maps in supervised learning. While saliency maps may help identify relevant features (e.g., pixels)  in the input for classification tasks of deep neural networks, similar frameworks are understudied in unsupervised learning. Therefore, we introduce a new method of obtaining saliency maps for latent representations of known or novel high-level concepts, often called concept vectors in generative models. \textit{Concept scores}, analogous to class scores in classification tasks, are defined as dot products between concept vectors and encoded input data, which can be readily used to compute the gradients. The resulting \textit{concept saliency maps} are shown to highlight input features deemed important for high-level concepts. Our method is applied to the VAE's latent space of CelebA dataset in which known attributes such as ``smiles'' and ``hats'' are used to elucidate relevant facial features. Furthermore, our application to spatial transcriptomic (ST) data of a mouse olfactory bulb demonstrates the potential of latent representations of morphological layers and molecular features in advancing our understanding of complex biological systems. By extending the popular method of saliency maps to generative models, the proposed concept saliency maps help improve interpretability of latent variable models in deep learning. \\

Codes to reproduce and to implement concept saliency maps: \underline{\url{https://github.com/lenbrocki/concept-saliency-maps}} \\
\end{abstract}

\section{Introduction}
A rapidly increasing amount of unlabeled data, such as images and molecular data, has prompted a rise of deep generative models, that can be trained without human supervision. By using a vast amount of unlabeled data, unsupervised learning models such as variational autoencoders (VAEs) \cite{Kingma2014, Rezende2014} extract low-dimensional latent spaces that compactly encode high-dimensional input data and potentially reveal hidden relationships. While deep generative models are capable of generating new images \cite{Salimans2015, kulkarni2015deep, mescheder2017adversarial} and enable manipulation of image-specific attributes \cite{Larsen2016, higgins2017beta}, it remains a grand challenge to achieve intelligible understanding of their behavior \cite{lipton2016mythos, kim2017interpretability, Adebayo2018}. We are interested in understanding and interpreting the latent representations of high-level concepts in generative models. Using VAEs, we propose to evaluate the importance of input features with respect to concept vectors and provide a new method of obtaining \textit{concept saliency maps}. This essentially extends the popular method of saliency maps in classification tasks to unsupervised learning. 

In predicting known classes using convolutional neural networks (CNNs), saliency maps have been introduced as a natural approach to make models interpretable \cite{erhan2009visualizing, baehrens2010explain, Simonyan2014}. A saliency map visualizes relative importances of the input pixels with respect to the classes that the neural network has been designed and trained on. It gives an insight into the behavior of the model that leads to a certain prediction. Saliency maps are obtained by calculating the gradient of the class score $S_
\textnormal{class}$ with respect to the input pixels $p_{ij}$, where $S_
\textnormal{class}$ is usually taken to be the activation of the neuron in the output layer encoding the class of interest. They attempt to answer the question: ``Which pixels were decisive for this particular classification made by the model?" The originally proposed method of obtaining the gradients may give noisy saliency maps, which prompted a number of improvements and variants on the calculation and backpropagation of the gradients \cite{Springenberg2015, Smilkov2017SmoothGradRN, sundararajan2017axiomatic}.
\begin{figure}[!t]
	\centering
	\includegraphics[width=0.5\textwidth]{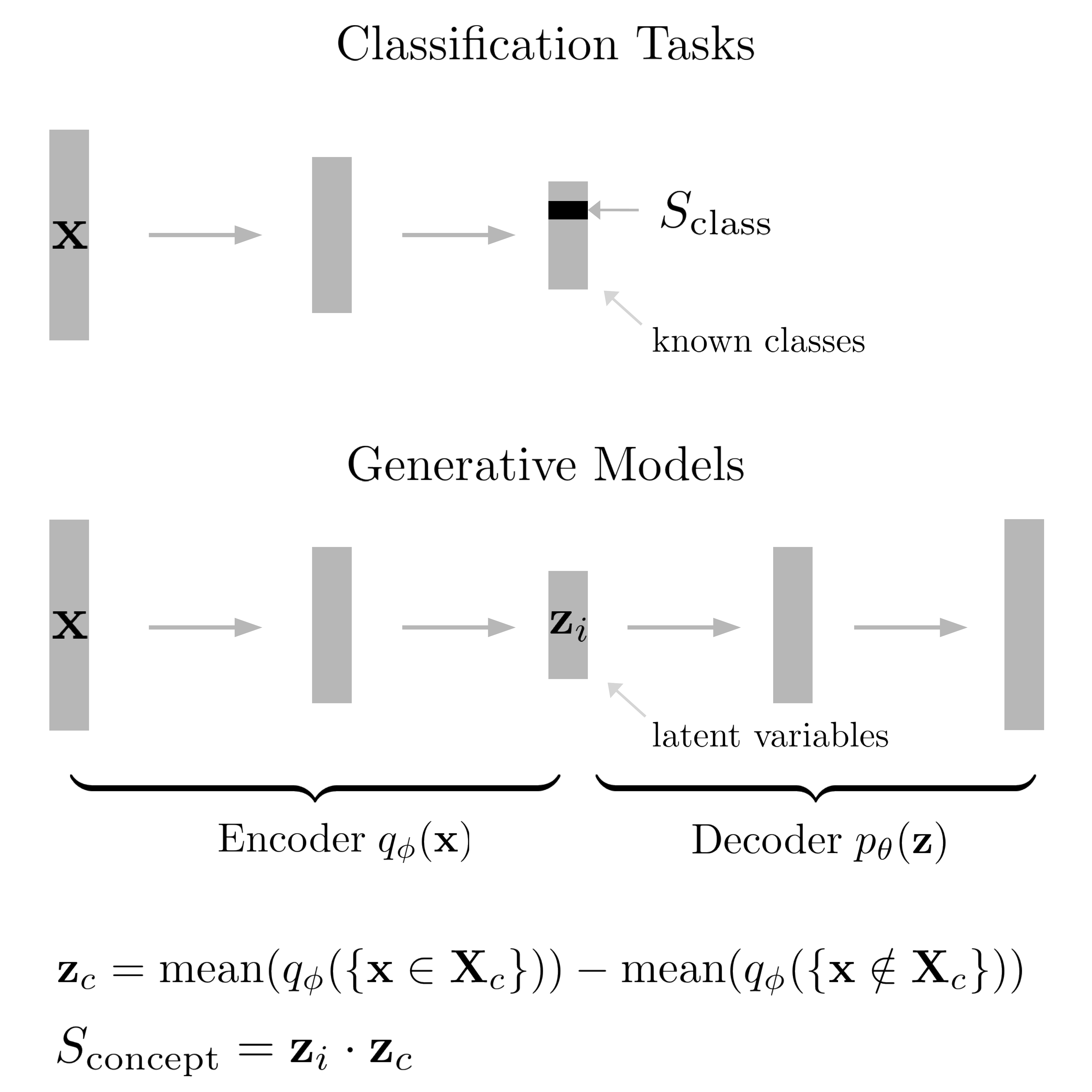}
	\caption{\small An overview of the proposed method of obtaining concept saliency maps. Top: In a classification task, the activation of the output neuron encoding the class of interest serves as class score $S_
\textnormal{class}$ to obtain saliency maps. Bottom: A deep generative model, such as VAE encodes input $\bx_i$ in the latent space and reconstructs the input from $\bz_i$. The concept score $S_\textnormal{concept}$ is defined as the dot product of the latent representation of the input image $\mathbf z_i$ and the concept vector $\mathbf z_c$, where $\bX_c$ is the set of images expressing a certain concept. }
	\label{fig:overview}
\end{figure}
This work aims to generalize the method of saliency maps to be applicable in generative models, particularly in deep latent variable models such as variational autoencoders (VAEs). VAEs are among the most popular approaches in unsupervised learning of complex distributions \cite{Kingma2014, Rezende2014}. They have been demonstrated to be capable of generating complicated imagery such as handwritten digits \cite{Kingma2014, Salimans2015}, faces \cite{kulkarni2015deep, mescheder2017adversarial}, and others. Furthermore, it has been shown that VAEs learn a meaningful latent representation which allows the manipulation of attributes through traversal in latent space \cite{Larsen2016, higgins2017beta}. 

\begin{figure}[!t]
	\centering
	\includegraphics[width=0.5\textwidth]{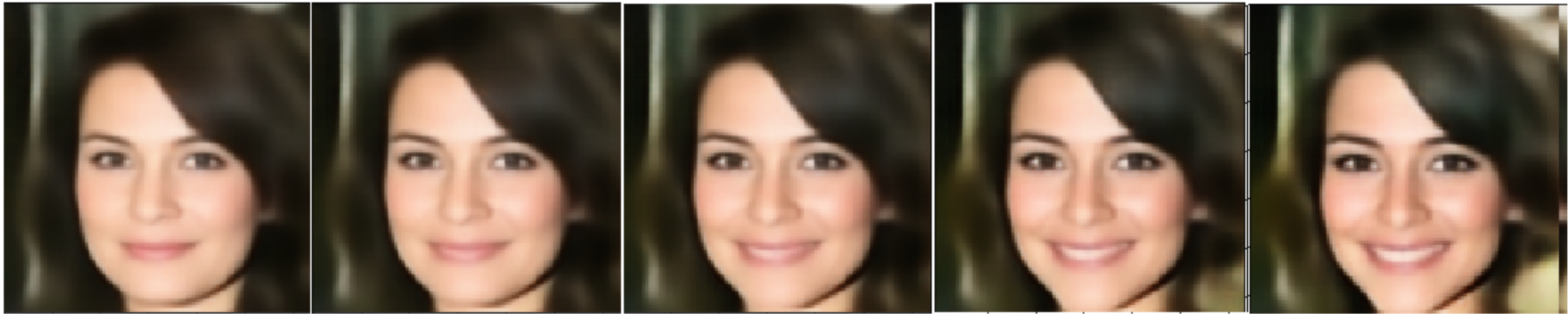}
	\caption{\small An example of meaningful concept vectors. Known attributes `smile' is used to construct a concept vector using a VAE of the CelebA data. Adding this smile concept vector to a latent representation of a given picture changes the facial features associated with a smile while leaving other features mostly unchanged. }
	\label{fig:smile}
\end{figure}

To achieve applicability to VAE and other generative models, which naturally lack known classes, we propose to use concept vectors to compute concept scores $S_\textnormal{concept}$, which can be understood as a replacement for $S_
\textnormal{class}$ (Fig. \ref{fig:overview}). A concept vector is a latent representation of a high-level concept, which could be known attributes \cite{Larsen2016, White2016}, cluster memberships \cite{Huang2016}, or others. Such a concept vector has been demonstrated to be capable of manipulating an image by adding a certain attribute, as is demonstrated in Fig. \ref{fig:smile} \cite{Larsen2016, White2016}. Once a generative model is trained on some dataset $\bX=\{\bx_i\}_{i=1}^N$, a concept vector $\mathbf z_c$ is readily obtained by averaging over the latent representations of samples containing an attribute of interest and subtracting the average of samples which do not. The concept score $S_\textnormal{concept}$ is then obtained by measuring the similarity of the latent representation of an input image $\bz_i = q_{\phi}(\bx_i)$ and the concept vector corresponding to that attribute $\bz_c$, where $q_{\phi}(\bx_i)$ is the encoder of the VAE.

Our proposed method obtains input-specific saliency maps for generative models by weighing the input pixels by relevance with respect to this concept vector. In other words, we answer the question: ``Which parts of a given image are particularly relevant for this concept?". This \textit{concept saliency map} is obtained by calculating the dot product between the latent representation of a certain image and the concept vector, although a different method to calculate the concept score may be useful in other domains. The concept score $S_\textnormal{concept}$ is motivated by the intuition that two vectors with a high dot product are more aligned, and thus more similar, than ones with a low dot product. Therefore, it can be seen as analogous to a neuron in the prediction vector of a classifier.

In this sense our proposed method is a generalization of saliency maps to generative models and is not limited to a classifier's possible predictions. The user is free to utilize known attributes or construct novel concepts based on the data. Utility of the concept saliency maps are demonstrated through application to a large-scale face image database, CelebA \cite{liu2015faceattributes} and spatial transcriptomic (ST) data of a mouse olfactory bulb \cite{staahl2016visualization}.

\section{Related Work}

\subsection{Concept vectors}

Generative models, such as VAE, often operate on the assumption that there is a meaningful low-dimensional latent space, meaningful in the sense that it encodes high-level concepts. Identifying, estimating, and disentangling such latent space has become important due to their potential to explain how unsupervised learning works \cite{Reed2015, Larsen2016, White2016, kim2017interpretability}. Sampling from latent space may allow us to reconstruct a meaningful output and to approximate visual analogies \cite{Reed2015}. In \cite{Larsen2016}, a latent representation of a certain attribute, such as a smile, is obtained and added to an arbitrary image which was referred to as \textit{visual attribute vectors} (e.g., Fig. \ref{fig:smile}). They noticed a strong correlation among certain attributes such as ``heavy makeup'' and ``wearing lipstick''. Reference \cite{White2016} attempted to decouple correlated attributes when a correlation stems from a sampling bias. Note that in \cite{White2016}, the dot product of a latent representation of an image with a concept vector was used to build an effective binary classifier, which motivated our definition of the concept score $S_{\text{concept}}$.

High-level concepts in latent space can be further discovered in semi- or unsupervised manner. Reference \cite{Hong2015} uses a limited reference database of images with attributes to cluster and annotate a set of unlabeled input images. By coupling a CNN with a set of data-dependent binary attributes, \cite{Huang2016} seeks to automatically discover image attributes. Internal states of a deep neural network may be interpreted in terms of high-level concepts, which \cite{kim2017interpretability} called \textit{concept activation vector}. This enables one to determine how important a certain internal concept, such as stripes, is for the prediction of a class, for example zebra. We are interested in both known and unknown attributes which may be learned from the input data. The novelty of our approach is twofold: firstly, the applicability to latent variable models leveraging low-dimensional latent space and secondly, the usage of the dot products obtained from concept vectors to create saliency maps. 

\subsection{Saliency maps}

In supervised learning there exist many methods to attribute the prediction of a network to its input features. Saliency maps are defined as the gradient of the class score $S_\textnormal{class}$ with respect to the input pixels $p_{ij}$ \cite{erhan2009visualizing, baehrens2010explain, Simonyan2014}. Despite early successes, direct calculation of the gradients often leads to noisy saliency maps without clearly focused regions. Several propositions have been made to improve them, such as Guided Backropagation (GuidedBP) \cite{Springenberg2015}, Rectified Gradient (RectGrad) \cite{Kim2019} and SmoothGrad \cite{Smilkov2017SmoothGradRN}. The first two methods modify the back-propagation of the gradient through the Rectified Linear Activation Unit (ReLU) $f(x) = \text{max}(x,0)$, for a detailed description see sec. \ref{sec: grads}. SmoothGrad seeks to denoise the saliency map of a given image by sampling similar images by adding noise and averaging over the saliency of the sampled images. Integrated Gradients \cite{sundararajan2017axiomatic} computes interpolations between a given image and a baseline image and integrates the saliency maps of these interpolated images which essentially alleviates the sensitivity to the  saturation of the input pixels. Other methods such as Layer-wise Relevance Propagation (LRP)\cite{bach2015pixel, samek2016evaluating} and DeepLift \cite{shrikumar2017learning} are relevance score based techniques, which means that they propagate back the relevance score(which is equal to $S_\textnormal{class}$ in the final layer) via the activations of the previous layers without using gradients.

Our work essentially extends this family of methods to embrace unsupervised learning in which the class score is no longer available. We propose to extract the latent layer of VAE and related generative models and feed the dot products as the class score into any of the aforementioned methods. 

\section{Method}
In the context of supervised classifiers, conventional approaches to obtaining saliency maps calculate the gradient of the output neuron encoding the class of interest, i.e. the class score $S_
\textnormal{class}$, with respect to the input pixels $p_{ij}$. The gradient tells us which inputs need to be changed the least to have the biggest influence on the class score, essentially identifying the most significant inputs. To find the input pixels which are most significant in maximizing the class score one can simply clip all negative gradients. In unsupervised models we do not have a prediction vector to choose our class score from but instead a low dimensional latent space which reflects the input data and it is a priori not clear how to construct a saliency map from it.

Recent developments of variational Bayesian approaches resulted in variational autoencoder (VAE) \cite{Kingma2014, Rezende2014} and related methods that estimate meaningful latent spaces. Briefly, the general architecture of VAE consists of the encoder and the decoder, both of which consist of multiple layers of neural networks. The encoder, which typically performs drastic dimension reduction, compresses the observed data (the input) into the latent variables, while the decoder reconstructs the observed data from the latent variables (Fig. \ref{fig:overview}). The input data $\bX=\{\bx_i\}_{i=1}^N$ are generated by some latent variables $\bz$ with a prior distribution $p_{\theta}(\bz)$. Then, $\bX$ is realized from a conditional distribution $p_{\theta}(\bx | \bz)$, where $\theta$ and $\bz$ are unknown. We attempt to do inference in this model such that given $\bX$, find $\bz$, by calculating the posterior density
$p_{\theta}(\bz|\bx) = \frac{p_{\theta}(\bx|\bz)p_{\theta}(\bz)}{p_{\theta}(\bx)}$,
which is intractable since we would have to integrate over all latent variables $p_{\theta}(\bx) = \int d\bz\; p_{\theta}(\bx|\bz)p_{\theta}(\bz)d\bz$.

VAE circumvents this challenge by fitting an approximate inference model $q_{\phi}(\bz|\bx)$ which is an approximation to the true distribution $p_{\theta}(\bz|\bx)$. An iterative learning process jointly learns the parameters $\theta$ and $\phi$. The probabilistic encoder $q_{\phi}(\bz|\bx)$ and decoder $p_{\theta}(\bx|\bz)$ are constructed of neural networks with $L$ layers. In images and other spatial data, convolutional neural networks are often used to take advantage of spatiality \cite{Fukushima1980, Lecun1998}. To train the network, weights are adjusted through gradient descents to minimize a pre-specified loss function.

\subsection{Extracting concept saliency maps} \label{algorithm}

The latent space is thought to encode high-level concepts, such as facial attributes or morphological structures. Many of unsupervised learning techniques aim to disentangle this latent space, extract useful latent representations of high-level concepts, and present how they are manifested in the observed (input) data. Larsen \cite{Larsen2016} and others have shown that we can find directions, or concept vectors, in latent space, which encode a certain attribute. Instead of a class score, for generative models we propose to use the dot product of $\bz_i = q_{\phi}(\bx_i)$ and a certain concept vector $\bz_c$. We refer to this dot product between $\bz_i$ and $\bz_c$ as \textit{concept score}.

Our method to obtain saliency maps from concept vectors is summarized in the following algorithm. \\

\noindent \underline{Algorithm 1. Concept Saliency Maps}
\begin{enumerate}
	\item Train VAE on the observed data $\bX=\{\bx_i\}_{i=1}^N$
	\item Obtain the set of latent representations $\bZ = \{\bz_i\}_{i=1}^n$ by applying the encoder $q_{\phi}$ on $\bX$
	\item Identify the data attribute of interest (e.g. `smiling') and denote those samples with the data attribute of interest by $\bZ^+$ and without it by $\bZ^-$ 
	\item Obtain concept vectors $\bz_c$ as follows, where $n^+$ and $n^-$ refer to the number of samples with and without a certain attribute \label{eq:concept}
	\begin{align}
	\bz_c = \frac{1}{n^+}\sum{\bZ^+} - \frac{1}{n^-}\sum{\bZ^-}
	\end{align}
	\item Obtain the concept score $S_\textnormal{concept}$ by calculating the dot product between $\bz_c$ and $\bz_i = q_{\phi}(\bx_i)$ 
	\begin{equation}
	S_\textnormal{concept} = \bz_c\cdot \bz_i
	\end{equation}
	\item Calculate the gradient of the concept score with respect to input pixels $p_{ij}$ to generate saliency map $M_{ij}$
	\begin{equation}
	M_{ij} = \frac{\partial S_\textnormal{concept}}{\partial p_{ij}}
	\end{equation}
\end{enumerate}\normalsize

When the computationally intensive Step 1 of training VAE is previously completed, the trained model can be used to compute the concept scores of interest. A concept can be defined by annotations of the dataset or be user defined. In the latter case the user can choose a subset of images that represent a certain concept and obtain a concept vector by averaging as is explained in Step 4. Another way is to cluster the data, possibly in latent space, and use cluster membership as concept or, as is exemplified in Sec. \ref{sec: ST}, a concept can be formed by highly correlated samples. It is data dependent which method is most suitable.

While we have focused on the dot product in the Step 5, it is possible to adapt modifications and improvements to how the concept score is derived. Dependent on data types and analysis goals, one may explore cosine similarity coefficients, $L_p$ norms and others and assess how well the input data with and without certain attributes are separated. The last step of computing the gradients, which is an active area of research, is explained in detail below.

Our method can readily be used to explore and visualize the latent space in pre-trained VAE models, as long as one is able to extract the latent representation of the input samples.

\subsection{Calculating the gradients}\label{sec: grads}

There are different methods of calculating the gradients which are used to obtain saliency maps. They differ in how they handle the backpropagation of the gradient through the rectified linear activation unit (ReLU), which is used throughout in our neural networks and is defined as $f(x) = \text{max}(x,0)$. Suppose we have a $L$-layer densely connected network and denote the input of neuron $i$ in layer $l$ as $x^i_l$ and the weight between neurons as $w_{ij}$. By larger $n$ we denote ``deeper'' layers, such that $n=0$ is the input layer and $l=L$ the output layer. The output layer in a classification task would contain the class score, whereas in the generative models, we use the proposed concept score. The gradient $R^i_l$ is defined as
$R^i_l = \frac{\partial S_{\text{concept}}}{\partial x^i_l}$
and tells us how the class score or the concept score changes with respect to $x^i_l$, where $x^i_l$ could for example be the values of the input pixels. Using the chain rule one finds the following relation for the backpropagation of gradients:
\begin{equation}
R_l^k = \sum_{j}w_{kj}\mathbf{1}(x^k_l>0)\cdot R^j_{l+1}\;,
\end{equation}
where $\mathbf 1$ is the indicator function. 
The activation maps created using this backpropagation are known to be very noisy and therefore improved methods have been proposed. \textit{Guided Backpropagation} \cite{Springenberg2015} further demands that the gradients of the higher layer have to be positive in order to be propagated:
\begin{equation}
R_l^k = \sum_{j}w_{kj}\mathbf{1}(x^k_l\cdot R^j_{l+1}>0)\cdot R^j_{l+1}\;,
\end{equation}
and this additional guiding of the gradient leads to sharper activation maps.
Recently, \textit{Rectified Gradient} \cite{Kim2019} has been proposed as yet another method to calculate gradients. It introduces an external parameter $\tau$ which acts as a threshold for gradients to be backpropagated:
\begin{equation}
R_l^k = \sum_{j}w_{kj}\mathbf{1}(x^k_l\cdot R^j_{l+1}>\tau)\cdot R^j_{l+1}\;.
\end{equation}

Our proposed framework also works with methods of computing the gradients since it only replaces $S_{\text{class}}$ with $S_{\text{concept}}$ leaving anything else untouched. In the following applications, we present both Guided Backpropagation and Rectified Gradient which have their own advantages and disadvantages.

\begin{figure}[b!]
	\centering
	\includegraphics[width=0.5\textwidth]{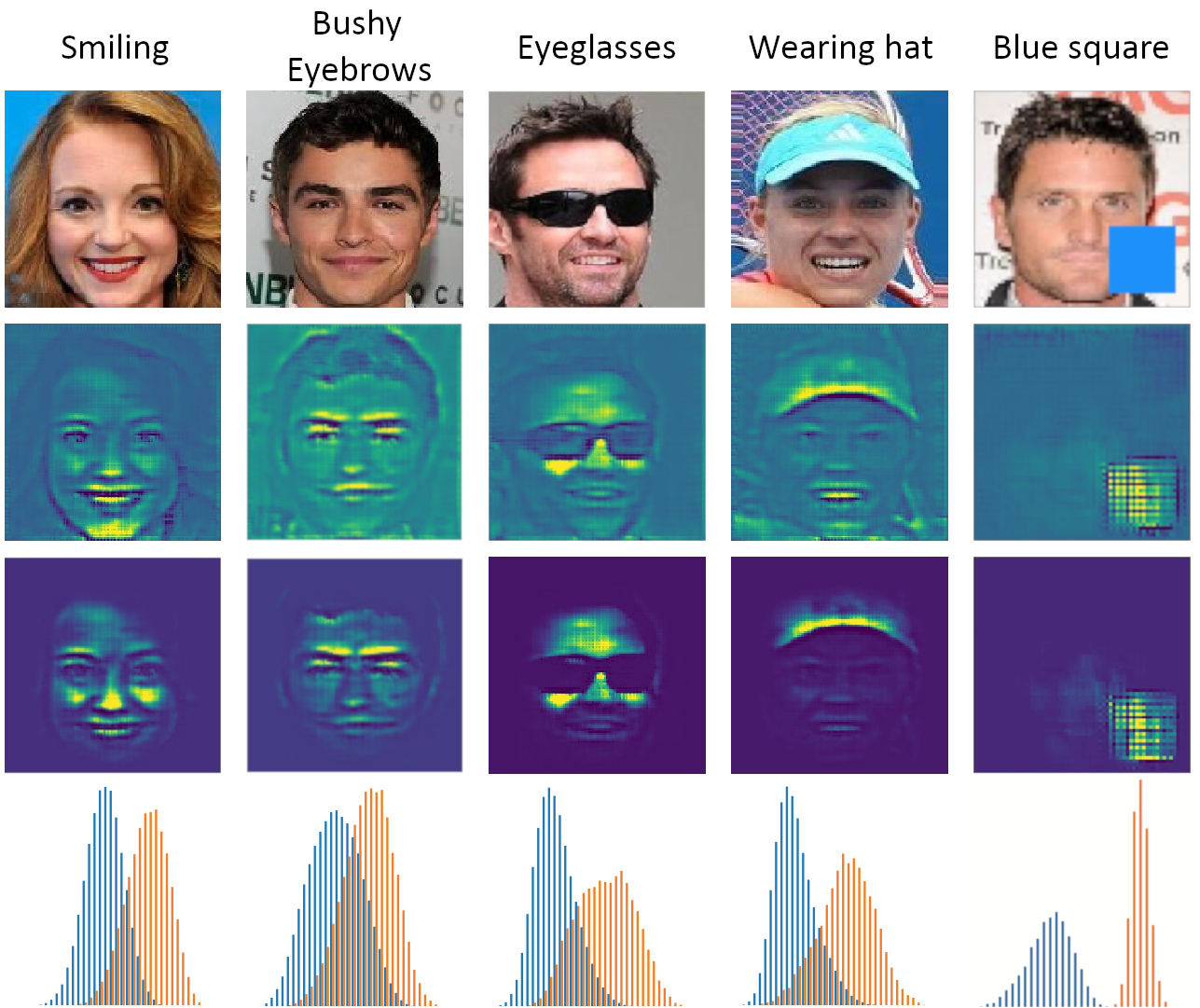}
	\caption{\small Saliency maps for selected facial attributes in the CelebA database. Concept vectors and concept scores are built from known attributes. For the example images with the highest concept scores, \textit{Guided Backpropagation}\cite{Springenberg2015} (second row) and \textit{Rectified Gradient}\cite{Kim2019} (third row) are used to obtain concept saliency maps. For the last column, artificial blue squares have been inserted into a subset of 10000 pictures and the VAE has been re-trained with this subset as part of the training data. The bottom row shows histograms of the dot products with coloring and axes as in Fig.  \ref{fig:compare}.}
	\label{fig:saliency}
\end{figure}

\section{Results}
We apply the proposed method of concept saliency maps to two datasets: a large-scale face database with attributes, CelebA, \cite{liu2015faceattributes} and a spatial transcriptomic (ST) dataset of a mouse olfactory bulb \cite{staahl2016visualization}. In the well-known CelebA, there exist annotations of facial features which can be used to visualize and evaluate. We demonstrate how known attributes such as smiles and glasses can be used to generate concept saliency maps. Spatial transcriptomics (ST) is a novel technique to measure gene expression profiles of a sample while maintaining spatial information. We use this spatial map of gene expression values located in a grid to demonstrate how to create saliency maps with limited prior knowledge about the input samples.

\subsection{CelebFaces Attributes Dataset (CelebA)}
CelebA is a large database of face images with known attributes \cite{liu2015faceattributes}, consisting of 202593 images and 40 binary annotations of facial attributes for each image. The images have been aligned, scaled and cropped to $128 \times 128$ pixels using the landmark annotations that come with the dataset \cite{liu2015faceattributes}. We used a VAE with convolutional layers, ReLu activation and batch normalization, down- and upsampling is done using strides and the dimension of the latent layer is 400. For the full architecture please refer to Tab. \ref{tab:vae}. The network has been trained for 50 epochs using the Adam optimizer \cite{kingma2014adam} with learning rate set to 0.001.

\begin{figure}[t]
	\centering
	\includegraphics[width=0.4\textwidth]{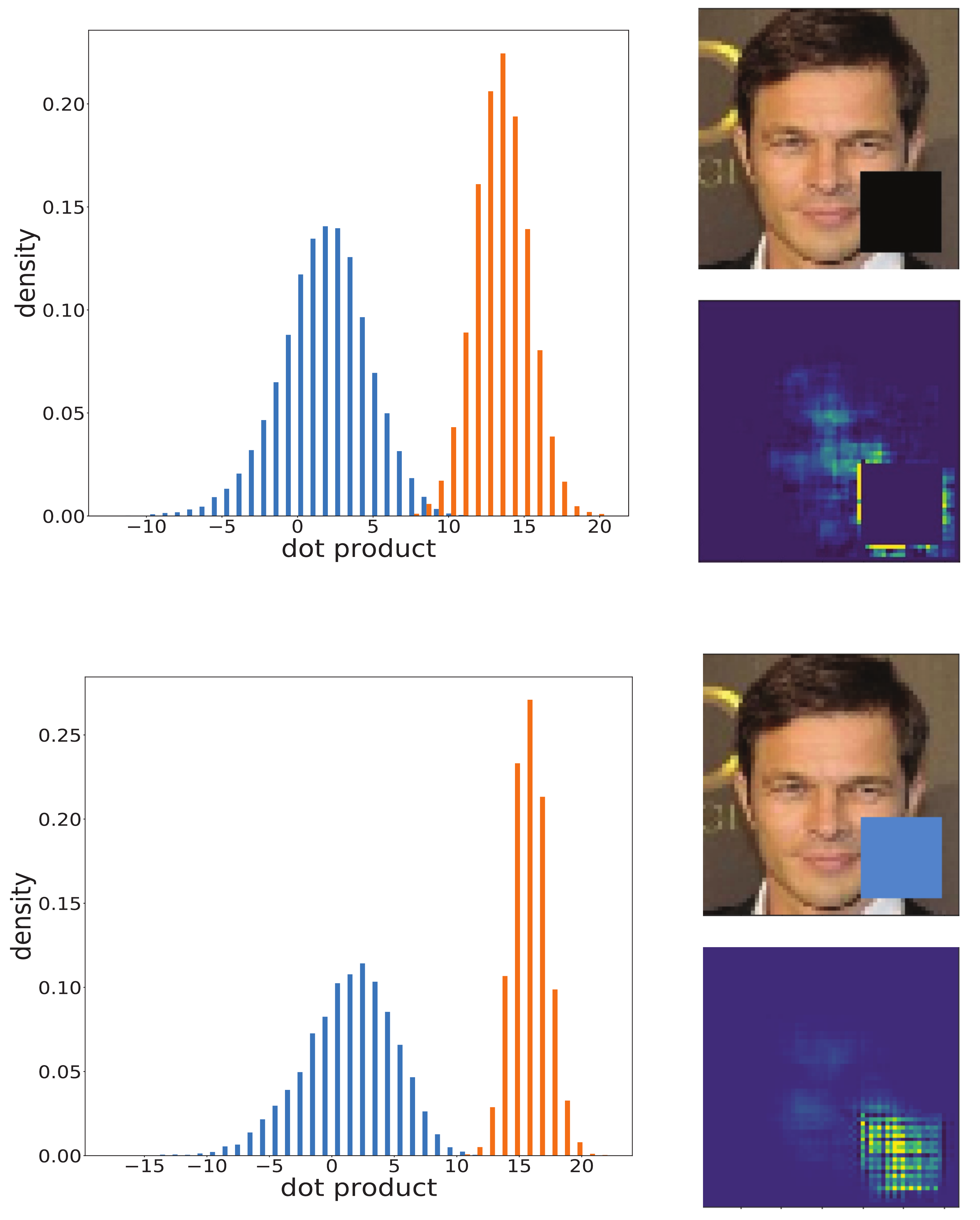}
	\caption{\small Impact of dark features on saliency maps. On the left, the histograms of the dot products of the concept vector for black (top) and blue (bottom) squares show that in both cases the distributions are well separated. However, the saliency map on the top fails to highlight the square, the attribute of our interest. This suggests that this failure can be attributed to the calculation of the gradients and not the concept score. The checkerboard pattern \cite{odena2016deconvolution} seen in the bottom map appears due to the strides in the convolutional layers. Orange bars correspond to images with a square and blue ones without.}
	\label{fig:compare}
\end{figure}

\begin{table*}[!t]
	\caption{\small Architecture of VAE for CelebA}
	\centering 
	\begin{tabular}{l l l l} 
		\toprule
		Encoder & Output size & Decoder & Output size \\ [0.5ex] 
		\midrule 
		Input image & $128\times 128  \times 3 $ & 400 fully-connected (latent layer) & 400 \\
		$5 \times 5$ 64 conv., stride 2, BN, ReLu & $64\times 64\times 64$ &  $1024\cdot 4\cdot 4$ fully-connected, BN, ReLu & 16384\\ 
		$5 \times 5$ 128 conv., stride 2, BN, ReLu & $32\times 32\times 128$ &$5\times 5$ 512 conv., stride 2, BN, ReLu & $8\times 8\times 512$  \\
		$5\times 5$ 256 conv., stride 2, BN, ReLu & $16\times 16\times 256$ &$5\times 5$ 256 conv., stride 2, BN, ReLu&$16\times 16\times 256$ \\
		$5\times 5$ 512 conv., stride 2, BN, ReLu & $8\times 8\times 512$&$5\times 5$ 128 conv., stride 2, BN, ReLu &$32\times 32\times 128$\\
		$5\times 5$ 1024 conv., stride 2, BN, ReLu & $4\times 4\times 1024$ &$5 \times 5$ 64 conv., stride 2, BN, ReLu&$64\times 64\times 64$  \\  
		512 fully-connected, BN, ReLu & 512&$5 \times 5$ 3 conv., stride 2, BN, ReLu & $128\times 128\times 3$ \\
		400 fully-connected (latent layer)& 400\\
		\bottomrule
	\end{tabular}

	\label{tab:vae}
	
\end{table*}

The saliency maps shown in Fig. \ref{fig:saliency} were obtained by employing the algorithm in section \ref{algorithm} using 2000 images for each attribute for averaging, where the annotations of the CelebA dataset for facial attributes such as ``smiling" or ``wearing hat" have been used. The saliency maps often match with the intuition that we have, they highlight eyebrows, the hat and the blue square clearly. For the smiling concept GuidedBP focuses rather on the mouth and chin region whereas RectGrad on teeth and cheeks. It is apparent that RectGrad produces cleaner maps than GuidedBP which is due to the additional threshold introduced. This threshold prevents RectGrad from performing a partial image recovery, which is the case for GuidedBP, as has recently been proven in \cite{nie2018theoretical}. 

However, note that not all attributes are created equal. First, there are continuous attributes -- e.g., ``rosy cheeks'' or ``oval face'' --  that are very challenging even for humans to agree upon. Second, some attributes -- e.g., ``attractive'' or ``young''  -- are highly subjective and do not necessary have common visual features. Third, there are correlated attributes -- e.g., ``heavy makeup'' and ``wearing lipstick'' -- whose underlying latent representations are intertwined. We have focused on attributes that are seemingly separated and well-represented. Fourth, we found that attributes that are inherently dark often are not visualized in saliency maps. For example, the eyeglasses are problematic because, although the correct region of the face is highlighted, the glasses themselves remain dark (Fig. \ref{fig:saliency}). This problem is connected to the calculation of the gradients which fails for very dark regions of the images. Reversely, bright regions tend to be overrepresented in the saliency maps as can be seen in the ``wearing hat" example, where the teeth and parts of the racket are also highlighted.

To further demonstrate that this is an effect due to the calculation of the gradients and not connected to the calculation of the concept score we have inserted also black squares into the pictures and calculated the dot products of the concept vector for blue and black square with the latent vectors of 20000 images in each case, where half of the images contained either a blue or black square and the remaining ones did not. Plotting the histograms shows that in both cases the dot product is significantly higher if a square is present but the saliency map fails to highlight the black square as the dominant feature (Fig. \ref{fig:compare}). In both cases the face almost vanishes, but in the black square case only the edges are partly highlighted.

\begin{figure}[!t]
	\centering
	\includegraphics[width=0.4\textwidth]{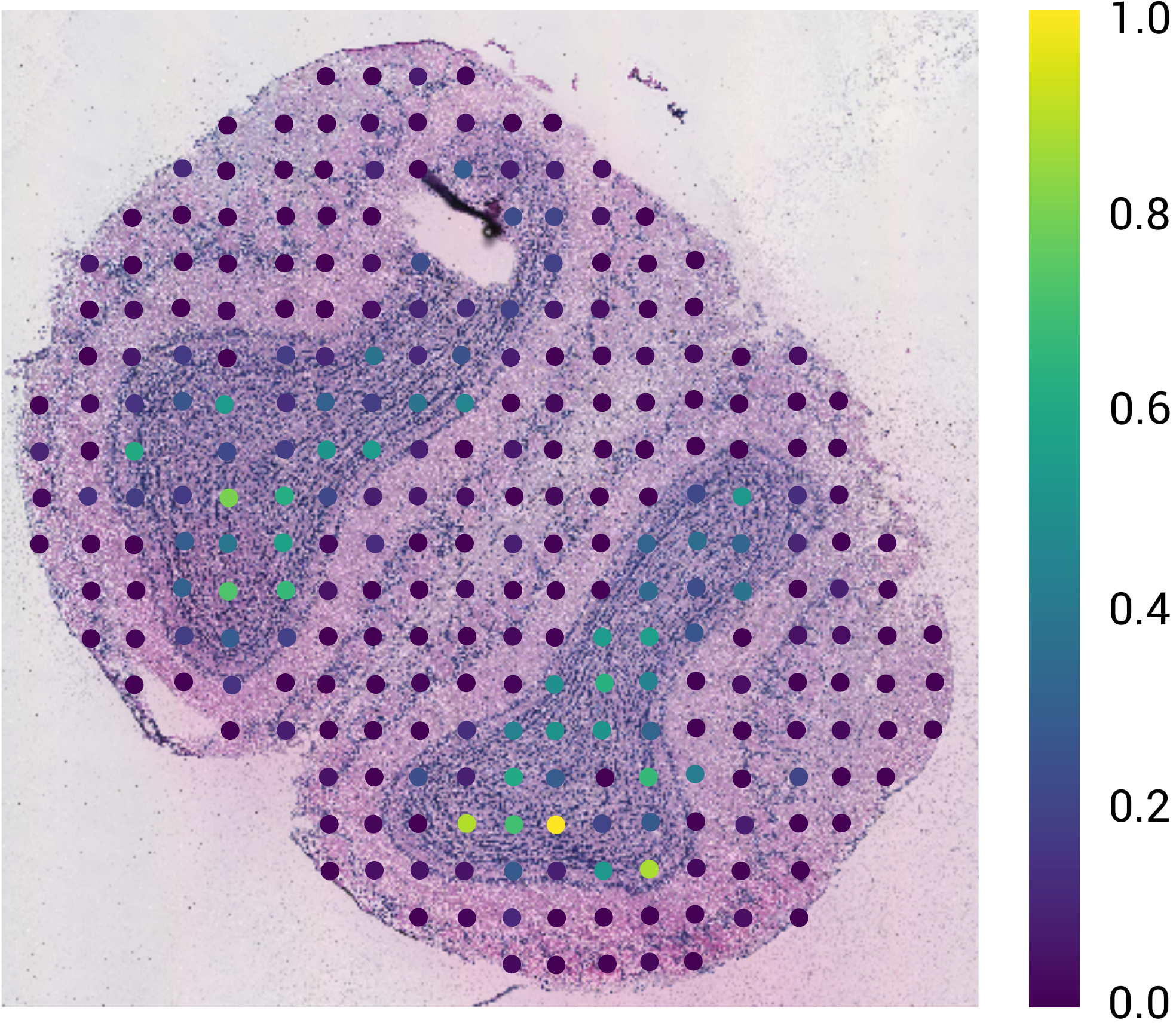}
	\caption{\small Visualization of spatial transcriptomics (ST) of a mouse olfactory bulb. ST provides gene expression profiles of a tissue section that are spatially resolved. The microscopic image \cite{staahl2016visualization} of the tissue section superimposed with the gene expression profile of \textit{Penk}. Colors represent normalized gene counts.}
	\label{fig:overlay}
\end{figure}

\subsection{Spatial Transcriptomics (ST) of an Olfactory Bulb}\label{sec: ST}
We are interested in understanding how high level concepts, such as morphological structures, are manifested on spatial gene expressions, which are measured by fixating and staining a sliced tissue sample. Particularly, the ST data of a mouse olfactory bulb contains genome-wide gene expressions from a sliced tissue section of a mouse olfactory bulb \cite{staahl2016visualization}. When positioned on a microchip, there are 267 spots in a grid that measure expression activities of up to 16573 genes within that locality. Each of 16573 genes is treated as a sample with 267 features, which are the counts of RNAs at the different spots in the tissue.

To provide a context for gene expression data in a spatially resolved tissue, a microscopic image of the tissue obtained by Hematoxylin-and-eosin staining has been superimposed with the normalized counts for the gene \textit{Penk} (Fig. \ref{fig:overlay}). Clearly, \textit{Penk} -- which is a proenkephalin gene playing a role in signaling receptor binding, response to stimulus, and cell projection -- is most highly expressed in the inner part of the tissue section known as the granular cell layer. We are interested in investigating the spatial expression of genes with known and unknown relations to morphological structures.

\begin{figure}[!t]
	\centering
	\includegraphics[width=0.4\textwidth]{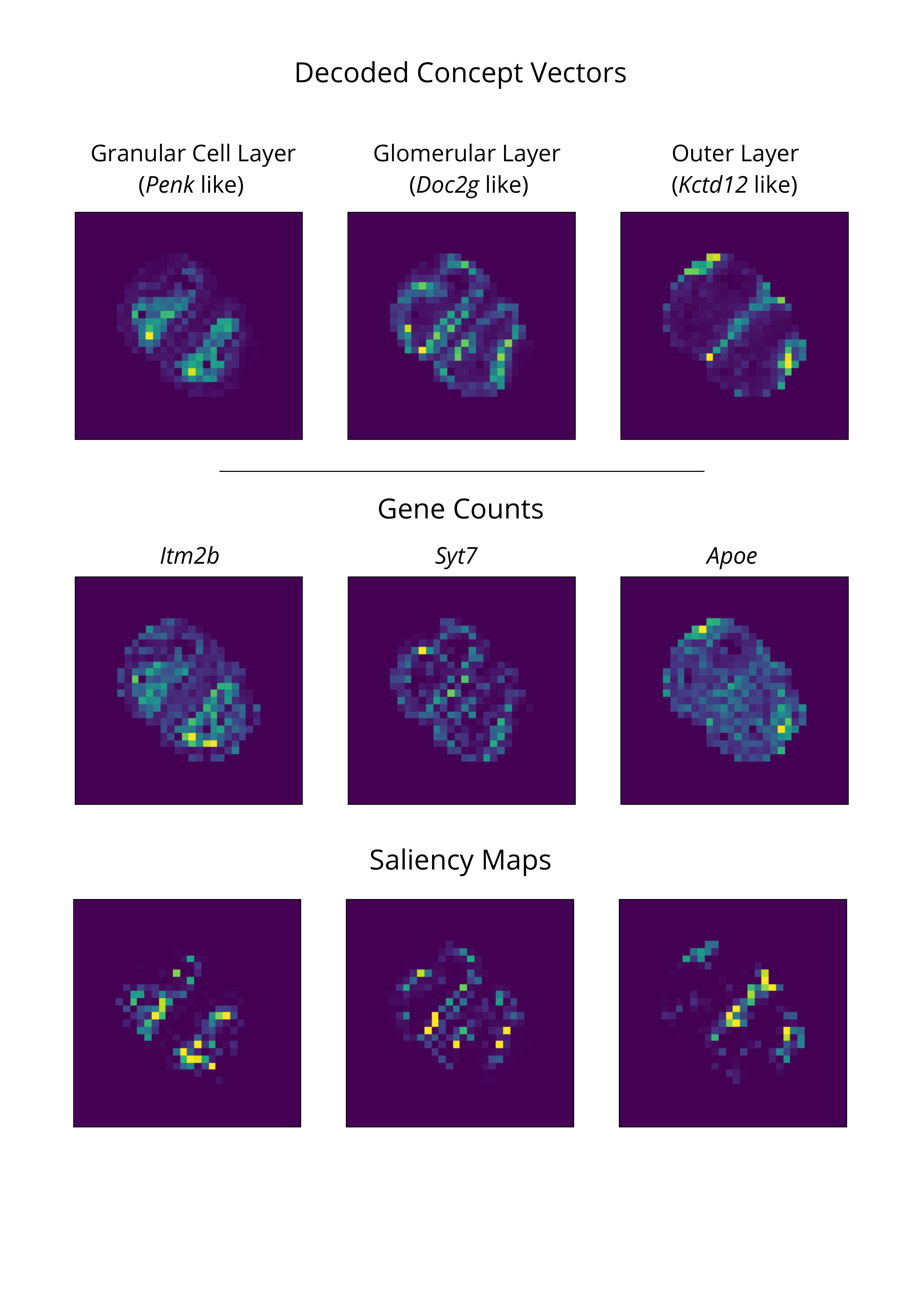}
	\caption{\small Concept vectors and saliency maps related to morphological layers in ST data. Top row: the decoded concept vectors for three morphological layers (derived from \textit{Penk}, \textit{Doc2g} and \textit{Kctd12}) are shown in $32\times 32$ matrices. Middle row: Gene count matrices for \textit{Itm2b}, \textit{Syt7} and \textit{Apoe}. Bottom row: Saliency maps for for \textit{Itm2b}, \textit{Syt7} and \textit{Apoe} with respect to the concept vectors on the top row. A color bar as in Fig. \ref{fig:overlay}.}
	\label{fig:STsaliency}
\end{figure}

To prepare the data for VAE, the gene counts have been arranged in $32\times 32$ matrices and normalized for each gene separately (see middle row in Fig. \ref{fig:STsaliency}). The VAE architecture used for this ST dataset is similar to the one used in CelebA, but with fewer layers and filters; for a detailed description see Tab. \ref{tab:st}. 

\begin{table*}[!t]
	\caption{\small Architecture of VAE for spatial transcriptomics dataset}
	\centering 
	\begin{tabular}{l l l l} 
		\toprule
		Encoder & Output size & Decoder & Output size \\ [0.5ex] 
		\midrule 
		Input image & $32\times 32  \times 1 $ & 20 fully-connected (latent layer) & 20 \\
		$4 \times 4 $ 16  conv., stride 2, BN, ReLu & $16\times 16\times 16$ &  $1024$ fully-connected, BN, ReLu & 1024\\ 
		$4 \times 4 $ 32 conv., stride 2, BN, ReLu & $8\times 8\times 32$ &$4\times 4$ 32  conv., stride 2, BN, ReLu & $8\times 8\times 32$  \\
		$4 \times 4 $ 64 conv., stride 2, BN, ReLu & $4\times 4\times 64$ &$4\times 4$ 16 conv., stride 2, BN, ReLu&$16\times 16\times 16$ \\
		256 fully-connected, BN, ReLu & 512&$4 \times 4$ 1 conv., stride 2, BN, ReLu & $32\times 32\times 1$ \\
		20 fully-connected (latent layer)& 400\\
		\bottomrule
	\end{tabular}
	
	\label{tab:st}
	
\end{table*}
 
In contrast to CelebA, this ST data do not come with conventional attributes that could be used to form concept vectors. Nonetheless, the advantage and motivation of ST is that gene expression profiles at different locations in a tissue section are crucial for complex molecular systems. How are genes expressed differentially across a tissue section? Reference \cite{staahl2016visualization} has shown that some genes present clear spatial organizations such as \textit{Penk}, \textit{Doc2g} and \textit{Kctd12}. These genes are specifically over-expressed in certain regions -- \textit{Penk} in the granular cell layer (GCL), \textit{Doc2g} in the glomerular layer (GL) and \textit{Kctd12} in the outer layer (see Fig. 2(b) in \cite{staahl2016visualization}). Therefore, we use these known genes to identify concept vectors for 3 morphological layers. Particularly, we calculated Pearson correlations in the latent space between these three genes and all other genes. Concept vectors related to these morphological layers were approximated by averaging over 50 genes with the highest correlation statistics (Fig. \ref{fig:STsaliency}).

We applied the proposed methods using these concept vectors to available genes in the ST data. In Fig. \ref{fig:STsaliency} (bottom row) the saliency maps for the genes \textit{Itm2b}, \textit{Syt7} and \textit{Apoe} with respect to the concept vectors in the top row are shown. It appears that the saliency maps indeed highlight the spots of the gene count matrices which match with the corresponding morphological layers. For instance, the saliency map for \textit{Itm2b} focuses on the granular cell layer (inner parts) and lets the glomerular and olfactory nerve layer (outer layers) almost vanish. 

\section{Conclusion and Future Work}
Unsupervised deep learning is becoming more critical as we are accumulating a greater amount of unlabeled data. One of the main goals of unsupervised learning is to extract meaningful and useful latent representations of known and novel concepts. To this end, we have developed a method of calculating \textit{concept scores} and obtaining \textit{concept saliency maps} which help identifying important regions or features in input data. Briefly, the concept score is defined as the dot product between the latent representation of an input and a concept vector. This proposed concept score can be understood to be analogous to the activation of a neuron in the prediction vector of a classifier, and our method therefore generalizes the technique for obtaining saliency maps to generative models. 

The effectiveness of our method is demonstrated by utilizing the CelebA dataset. When using known attributes, the concept scores are shown to effectively distinguish samples with and without a certain attribute and the concept saliency maps highlight relevant facial features. However, we have observed and investigated how very dark regions in an image tend not to be highlighted, even though they are crucial to the concept. It has been demonstrated that this is an effect due to the calculation of the gradients and not connected to the proposed concept score. Therefore, it would be interesting to further investigate how darkness impacts saliency maps and develop an improved gradient method or normalization technique that would not be biased.

We have further presented a novel application of proposed methods to spatial transcriptomics (ST) of a mouse olfactory bulb \cite{staahl2016visualization}. Concept vectors have been formed by using genes which are highly correlated to three well-known genes whose spatial expression profiles are known to coincide with certain morphological layers. Concept vectors for different morphological layers have been obtained from genes with high correlation statistics relative to genes with known spatial expressions. Finally, the concept saliency maps highlight the regions in the spatial gene expressions which most strongly overlap with the morphological layers.

We plan to further develop the proposed methods using ST datasets in order to better understand how deep generative models could be used to disentangle the latent space of spatial gene expression. It would be instructive to explore functional annotations and other high-level concepts that are not directly related to morphology. These annotations may be used to reveal spatial characteristics for certain functions and saliency maps visualize the contribution of genes to these concepts. Furthermore, it would be interesting to explore other ST datasets, many of them related to cancer and other medical applications. Lastly, it will become more important to combine ST with RNA-sequencing and molecular experiments to validate the computational results.

Finally, saliency maps, which have primarily been used for interpretation of machine learning models, have the potential to become more useful and popular as an essential tool for exploratory data analysis. High-level concepts hidden in latent space of VAE and other generative models may be discovered through dimension reduction and clustering and saliency maps can be used to reveal the significant features. In particular, it would be important to extend this set of unsupervised deep learning approaches to non-image data for molecular and biomedical applications. 

\bibliographystyle{IEEEtran}
\bibliography{refs}

\end{document}